\documentclass{article}
\usepackage{ijcai19}

\usepackage{times}
\usepackage{soul}
\usepackage{url}
\usepackage[hidelinks]{hyperref}
\usepackage[utf8]{inputenc}
\usepackage[small]{caption}
\usepackage{graphicx}
\usepackage{amsmath}
\usepackage{booktabs}
\usepackage{algorithm}
\usepackage{algorithmic}
\usepackage{amssymb}
\usepackage{booktabs} 
\usepackage{multirow}
\usepackage{subcaption}
\title{Collaborative Metric Learning with Memory Network for Multi-Relational Recommender Systems}

\author{
	Xiao Zhou$^1$\and
	Danyang Liu$^2$\and
	Jianxun Lian$^3$\And
	Xing Xie$^3$\\
	\affiliations
	$^1$Department of Computer Science and Technology, University of Cambridge, UK\\
	$^2$University of Science and Technology of China, Hefei, China\\
	$^3$Microsoft Research, Beijing, China\\
	\emails
	xz331@cam.ac.uk,
	ldy591@mail.ustc.edu.cn,
	\{Jianxun.Lian, Xing.Xie\}@microsoft.com
}

\begin{document}

\maketitle

\begin{abstract}
  The success of recommender systems in modern online platforms is inseparable from the accurate capture of users' personal tastes. In everyday life, large amounts of user feedback data are created along with user-item online interactions in a variety of ways, such as browsing, purchasing, and sharing. These multiple types of user feedback provide us with tremendous opportunities to detect individuals' fine-grained preferences. Different from most existing recommender systems that rely on a single type of feedback, we advocate incorporating multiple types of user-item interactions for better recommendations. Based on the observation that the underlying spectrum of user preferences is reflected in various types of interactions with items and can be uncovered by latent relational learning in metric space, we propose a unified neural learning framework, named Multi-Relational Memory Network (MRMN). It can not only model fine-grained user-item relations but also enable us to discriminate between feedback types in terms of the strength and diversity of user preferences. Extensive experiments show that the proposed MRMN model outperforms competitive state-of-the-art algorithms in a wide range of scenarios, including e-commerce, local services, and job recommendations.
\end{abstract}

\section{Introduction}

With the ability to select the most relevant content from the flood of information resources, recommender systems have played an increasingly vital role in modern society by optimising user experiences for individuals and promoting business objectives for online platforms~\cite{tang2016empirical}. Keeping pace with the growing requirements of customisation and personalisation, recommender systems that are capable of learning fine-grained individual preferences with a concise, flexible, and efficient structure are eagerly expected.  

\begin{figure}
	\setlength{\abovecaptionskip}{0.1 cm}
	\setlength{\belowcaptionskip}{-12 pt}
	\centering
	\includegraphics[width = 8cm]{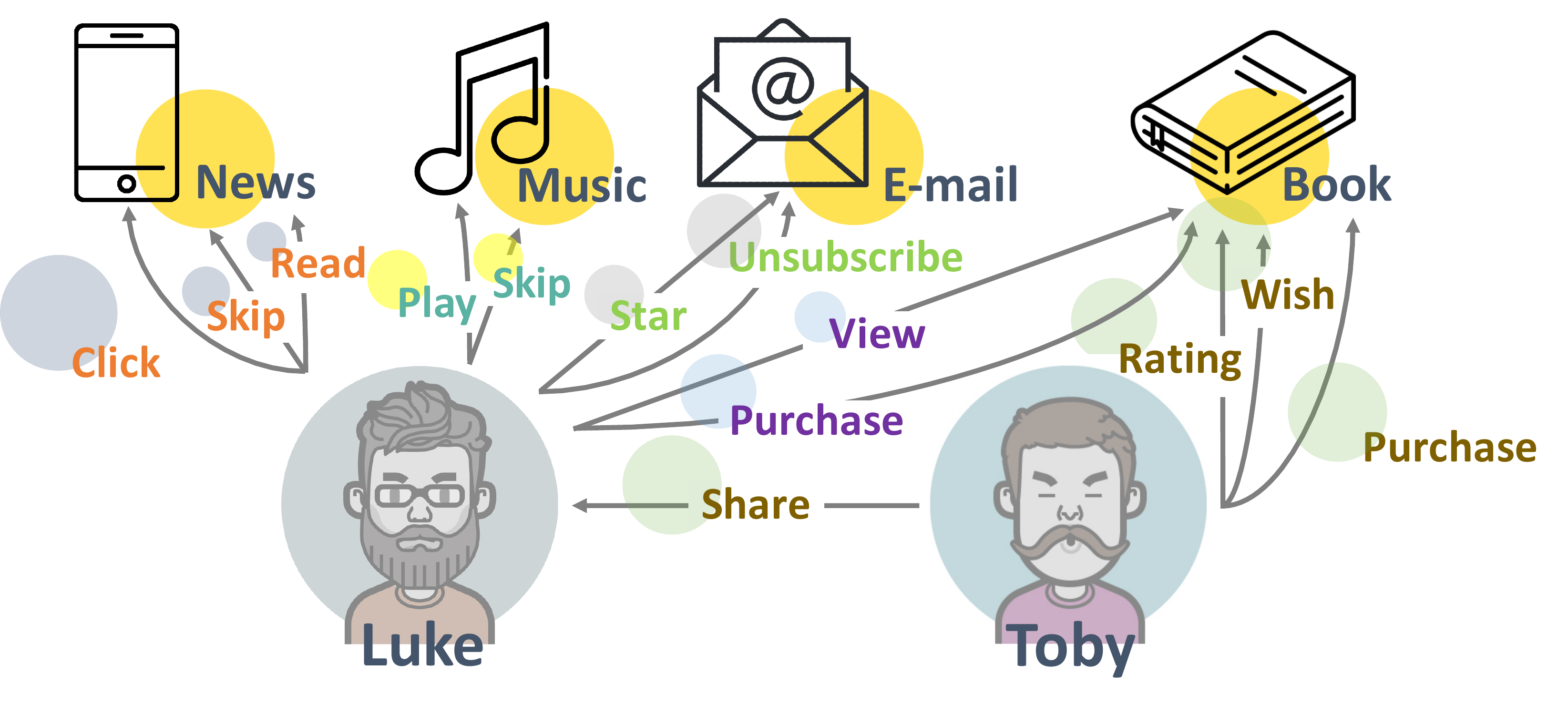}
	\caption{An example of multiple types of user feedback.}
	\label{fig:feedback}
\end{figure}

Given that mining valuable information hidden in users' historical interactions with items in multiple ways might hold the key to better characterise users, we devise a recommendation scheme that takes full advantage of various user feedback types in this paper. Before diving into the details, it is necessary to clarify the meaning of '\textit{multiple feedback types}', which can be varied in different scenarios. As exemplified in Figure~\ref{fig:feedback}, possible interaction types existing between users and recommended items include clicking, skipping, and reading for news recommendation; playing and skipping songs in music recommendation scenario; and purchasing, rating, and sharing for book recommendations. 

Most existing item-to-user recommender systems typically only employ one primary type of user-item interaction \cite{hu2008collaborative}, like clicking for online news recommendation and purchasing for the e-commerce scenario. Apart from primary feedback, online systems also allow users to leave additional types of feedback as listed above. Unfortunately, the power of using such extra feedback information to enhance recommendation performance has been largely neglected. In this paper, we leverage binary preference data in the implicit form and build a unified recommendation model that makes full use of multiple types of user-item interactions in various scenarios. Generally, it offers the following key advantages: 1) \textit{Alleviating data sparsity problem.} Compared to explicit data like numerical ratings, implicit feedback data are relatively cheap and widely available since they are usually gathered by systems automatically. Through utilising rich implicit feedback data, the sparsity issue can be effectively alleviated; 2) \textit{More precise and comprehensive user preference profile.} Incorporating multiple feedback types is essential for fine-grained detection of individual \textbf{preference diversity} and \textbf{strength}. Take the e-commerce scenario as an example, Toby likes Tesla's vehicles and always pays close attention to its latest models. However, he has never bought a Tesla in real life as the price is beyond his reach. In this case, if only purchasing feedback is taken into account, Toby’s interest in Tesla can hardly be captured. In another instance, when Luke considered buying a mobile phone, he browsed both iPhone 8 and iPhone XS before eventually buying the latter. By considering both browsing and purchasing feedback types, we can infer that Luke had a stronger preference for iPhone XS, relative to iPhone 8.

In this paper, we argue that jointly modeling multiple feedback types can help us reveal the underlying spectrum of user preferences in different dimensions and thus lead to better recommendation performance. More specifically, our ideas are materialised in the form of a neural learning framework that leverages the recent advancements of attention mechanism, augmented memory networks as well as metric learning. Our main contributions can be summarised as follows:


\begin{itemize}
\item We propose an end-to-end neural network architecture, named MRMN to address implicit collaborative filtering with multiple types of user feedback data.
	
\item External memory and attention mechanism are augmented in MRMN, making it capable of learning adaptive relational vectors in metric place for each specific user, item, and feedback type, and detecting fine-grained user-item relations and multidimensional preferences.
	
\item MRMN uncovers the underlying relationships between feedback types and shows multi-task ability to predict various types of user actions using one unified model.
	
\item Comprehensive experiments on real-world datasets demonstrate the effectiveness of MRMN against competitive baselines in various recommendation scenarios.
	
\item Qualitative analyses of the attention weights provide insights into the learning process of relations and illustrate the interpretability of MRMN in capturing higher order complex interactions between users and items.
	
\end{itemize}

\section{Related Work}
The past two decades have witnessed tremendous advances in the recommendation techniques from content-based filtering \cite{pazzani2007content} to collaborative filtering \cite{ekstrand2011collaborative}, from explicit feedback \cite{koren2008factorization} to implicit feedback \cite{he2016vbpr}, and from shallow models to deep models \cite{lian2018xdeepfm}. While most recommender systems typically take one type of user-item interaction into account, we emphasise the critical importance of incorporating multiple types of user feedback into the recommendation. A review of existing literature on recommender systems with multiple feedback types is provided below. 

Based on LinkedIn recommendation products, \cite{tang2016empirical} presented a general investigation of possible ways to incorporate multiple types of user feedback from an empirical standpoint. To make a more specific discussion, other studies considering multiple feedback types in recommender systems can be mainly divided into three categories: sampling-based, model-based, and loss-based approaches. For the first group of methods focusing on sampling refinements, \cite{loni2016bayesian} extended the sampling method of vanilla Bayesian Personalized Ranking (BPR) \cite{rendle2009bpr} to distinguish different strengths of user preferences reflected by various feedback types. Similarly, \cite{ding2018improved} leveraged view data in e-commerce recommendation and developed a view-enhanced sampler for classical BPR. Another line of works that sought to utilise multi-feedback interactions focused on model modification. For instance, \cite{liu2017personalized} developed the MFPR model that employed one type of explicit feedback (e.g., ratings) and several types of implicit user feedback (e.g., viewing, clicking logs) as input. Based on LSTM networks, \cite{li2018learning} designed an architecture that enables us to learn the short-term motivation and long-term preferences of users for the next purchase recommendation. 
By devising tailored loss function, \cite{ding2018improving} emphasised the importance of integrating view feedback in e-commerce recommendation and added pairwise ranking relations between purchased, viewed, and non-interacted actions instead of applying pointwise matrix factorization methods.

Through the literature review above, it can be found that the existing approaches proposed to handle multi-feedback recommendation tasks are relatively simple. There is a lack of in-depth investigation of relationships between user feedback types. In most cases, the importance weights of multiple feedback types were set manually. Moreover, current studies generally built the model in one particular recommendation scenario that a generic and adaptive architecture for multi-relational recommendation is still missing.

\section{Background}
\label{sec:background}
To fill the research gap discovered in the current literature, we build a novel neural network framework that can learn multi-relational vectors for each specific user, item, and feedback type adaptively and flexibly. Our work is highly inspired by recent advances in metric learning and memory networks, which are introduced in the following two subsections. 

\subsection{Metric Learning based Recommendation}
Over the last decade, matrix factorization (MF), as one of the most outstanding representative techniques of collaborative filtering (CF), has gained rapid acceptance in the field of recommender systems. Essentially, the main purpose of MF-based methods is to extract features of users and items by mapping them to a latent vector space. By doing this, MF can capture the existing user-item interactions approximately and infer the missing values by inner product for further recommendation \cite{tay2018latent}. Even though it is able to detect the most prominent features, we can hardly expect MF to uncover more fine-grained user preferences and to generate interpretable recommendations. From a fire-new perspective, \cite{hsieh2017collaborative} applied metric learning techniques to CF, and proposed collaborative metric learning (CML) algorithm. The essential distinction between CML and MF is that CML maps users and items in a metric space to minimise the distance between them if positive interactions exist. However, for many-to-many collaborative ranking task, fitting all positive user-item pairs into the same point in vector space would inevitably cause geometrical congestion and instability. To overcome this limitation, \cite{tay2018latent} proposed Latent Relational Metric Learning (LRML) to learn a latent relation vector for each given user-item pair. The architecture of our proposed model takes advantage of this metric-based learning scheme and enables multi-relational modeling in metric space for the collaborative ranking task, which obeys the crucial triangle inequality to capture fine-grained user preferences. 


\subsection{Memory Networks for Recommendation}

The memory network was initially proposed in the context of question answering \cite{Weston2015mem}. Until recently, it has been applied to recommender systems. \cite{huang2017mention} extended end-to-end memory networks to recommend hashtags for microblogs. \cite{chen2018sequential} employed memory networks for e-commerce recommendations using historical purchase records of users. \cite{ebesu2018collaborative} proposed a collaborative memory network, where memory slots were used to store user preferences and encode items' attributes. \cite{zhou2019topic} proposed a topic-enhanced memory network to optimise point-of-interest recommender systems.

The success of these recent applications highlight that the memory network architecture is efficient and flexible enough to perform joint task learning \cite{ebesu2018collaborative}, making it a suitable approach to our multi-relational learning task naturally. To the best of our knowledge, no prior work has employed the memory network for collaborative metric learning using multiple types of user feedback in recommender systems. Then, how to apply the memory mechanism properly in our case becomes a key focus. In question answering, the most classical application of memory networks, a short passage is provided along with a relevant question, the answer to which can be generated automatically via memory blocks. If we analogise our multi-relational recommendation task to a question answering problem, the question we are asking now becomes how likely a user would enjoy an item she has never interacted with, based on the user-item relations learned from the memory network module.

\section{Proposed Model} \label{proposed model}

\begin{figure}
	\setlength{\abovecaptionskip}{0.1 cm}
	\setlength{\belowcaptionskip}{-10 pt}
	\centering
	\includegraphics[width = 8.65cm]{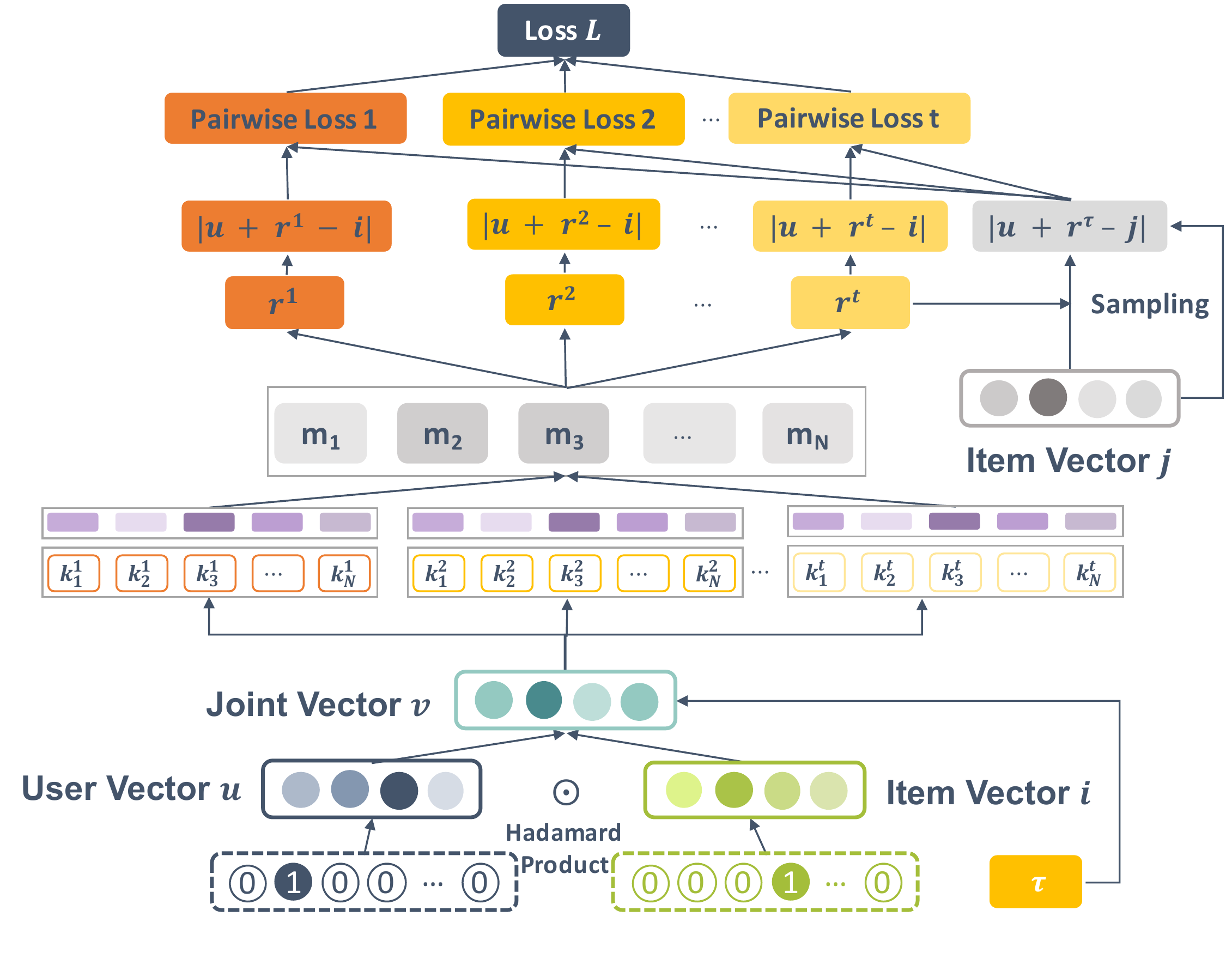}
	\caption{The architecture of MRMN.}
	\label{fig:architecture}
\end{figure}

We introduce a unified hybrid model called Multi-Relational Memory Network (MRMN) for personalised recommendation with multiple user-item interactions in this section. The overall architecture of MRMN model is presented in Figure \ref{fig:architecture}. Generally speaking, we fuse a memory component and neural attention mechanism to learn multi-relational vectors for each specific user-item pair in a non-linear fashion and produce the ranking score of candidate items to the target user for recommendation. In the following subsections, we will first give the formulation of our prediction task before introducing the structure of the proposed MRMN model in detail. 

\subsection{Problem Formulation}

Our research targets at recommender systems that operate multiple types of user feedback data in implicit form. Assume that we are given a set of users $U$ and a set of items $I$, associating with each other through $t$ types of implicit feedback ($t\geqslant 2$). For each feedback type $\tau$ in set $T$, the binary relations between $U$ and $I$ are represented by an interaction matrix $Y^{\tau} $, where an entry is set to 1 if the interaction exists between the pair of user and item. Otherwise, it is set to 0, meaning that the user has not interacted with the item yet. The main task of the model is to derive a prediction score for a given user-item pair, signalling the likelihood that the user would be interested in the item based on her historical interactions with items in multiple ways.

\subsection{User and Item Embedding}

For a training set consisting of triplets (user, item, $\tau$), the identities of a user and an item, represented as two binary sparse vectors via one-hot encoding, are employed as input features initially. Then the two vectors are projected to low-dimensional dense vectors to generate a pair of user and item embeddings, denoted as ($u$, $i$). After that, Hadamard product is applied to learn a joint user-item embedding $v$ through:
\begin{equation}
v = u\odot i
\end{equation}
The dimension of the generated vector $v \in \mathbb{R}^{d}$ is the same as $u$ and $i$, which is $d$. Apart from the pair of user and item, the input feedback type $\tau$ plays as a controller. Given that the MRMN is devised as a partially shared structure for various user feedback types, the observed interaction type $\tau$ for a certain user-item pair determines which part of the model would be activated during the training scheme. This particular design makes the learned relation vector specific to each triplet of a user, item, and interaction type, even though the embeddings of $u$ and $i$ are shared across various feedback types.

\subsection{Key Addressing and Attention Layers}

To discriminate the diverse interests of a user reflected via various interaction types, neural attention mechanism is used to read a memory matrix and generate multi-relational vectors for different feedback types adaptively. The main idea of the attention mechanism is to rescue the model from encoding all information into each memory slot. We assume that the multiple relations learnt from different feedback types for the same user-item pair can not only be distinguished from each other but also have inherent connections, characterised by the attention module over augmented memory blocks. 

Formally, a key matrix $ K^{\tau }\in \mathbb{R}^{ d\times N} $ is built for each feedback type $ \tau $, from which the attention vector of a particular feedback type can be obtained. Let $N$ represent the number of key slots in $ K^{\tau } $, which is a user-specified hyperparameter. For a given joint embedding vector $v$ and an observed interaction type $\tau$, the similarity between $v$ and each key slot $ k_{i}^{\tau } $ in the matrix $ K^{\tau }$ is calculated by dot product to obtain each element of the attention vector $ \omega^{\tau } $ for feedback $\tau$ by:

\begin{equation}
\omega_{i}^{\tau } = v^{T}k_{i}^{\tau } 
\end{equation}
Then the generated attention weights in vector $\omega^{\tau } $ are converted into a probability distribution over memory blocks using the softmax function to infer the importance of each memory slot's contribution for a given feedback type $\tau$.

\subsection{Multi-relational Vectors Generation}

Next the generated attention vector $\omega^{\tau } $ is utilised to learn a weighted sum of a sequence of memory slices in the memory matrix $M\in \mathbb{R}^{N\times d}$. Here the size of each memory slice is $d$, and the number of slices is $N$. The memory component is the core of the system where complex user-item relationships reflected in multiple feedback types are built. It is worth noting that different from the key addressing layer, where feedback types learn their key vectors, respectively; while here, they share the memory matrix. Each memory slice $ m_i\in \mathbb{R}^{d}$ in $M$ can be seen as a building block selected to form the relation vector for each feedback type according to its attention vector. In other word, the latent relation vector for feedback type $\tau$ denoted as $r^\tau$ is a weighted representation of memory matrix $M$, which can be generated as:
\begin{equation}
r^\tau = \sum_{i= 1}^{N}\omega_{i}^{\tau}m_i
\end{equation}

Through the architecture introduced above, a relation set $R$ consisting of multiple types of relation vectors for a specific user-item pair forms. 

\subsection{Optimisation}
\label{sec:opt}

Inspired by the multi-relational modeling technique of \textit{TransE} \cite{bordes2013translating} in knowledge base embedding, we define the scoring function of MRMN as:
\begin{equation}
s(u,i,\tau) = \left \| u+r^{\tau}-i \right \|_{2}^{2}
\end{equation}
where $\left \| \cdot \right \|_{2}$ denotes the $L_{2}$-norm. The $r^\tau$ employed to model the user-item relationship by a translation operating ensures the flexibility and superiority of MRMN in many-to-many recommendation (a user may enjoy many items, and many users can like an item) compared with the vanilla metric learning approach of CML, an ill-posed algebraic system mathematically which minimises the Euclidean distance between positive user-item pair via $\left \| u-i \right \|_{2}^{2}$.

We adopt the pairwise ranking loss for optimisation. Different from the pointwise loss \cite{hu2008collaborative} treating unobserved entries as negative samples, the basic idea of pairwise learning \cite{rendle2009bpr} is ranking observed entries higher than those unobserved. We also tried to combine point-wise loss and pair-wise loss in a unified framework, like that in~\cite{ding2018improving}. However, it turned out to be about 30\% worse than a margin-based approach. In this case, we do not integrate the point-wise loss in this paper. For each positive triplet ($u$, $i$, $\tau$) representing $u$ has an interaction of type $\tau$ with $i$, a corrupted triplet ($u$, $j$, $\tau$) is sampled according to the rule that $u$ has never interacted with item $j$. Mathematically, we define the objective function as:  


\begin{equation}
\mathcal{L}= \sum_{\tau}\sum_{\left ( u,i,\tau \right )\in \eta }\sum_{\left (u,j,\tau \right )\notin \eta } \phi (s(u,i,\tau) + \lambda ^{\tau}-s(u,j,\tau))
\end{equation}
where $\eta$ is the set of all positive triplets; $\lambda ^{\tau}$ is the margin separating the positive triplets and corrupted ones in terms of feedback $\tau$. The non-linear activation function  $\phi (\cdot)$ applied here is the rectified linear unit (\textit{ReLU}) function as we found it performed best empirically. Additionally, since MRMN is end-to-end differentiable, stochastic gradient descent (SGD) is employed to minimise the objective function.

It is worth mentioning that by injecting discriminative $\lambda ^{\tau}$, we can encode the \textit{strength} deviation corresponding to different feedback types. Here a larger setting of margin value will push the two points of positive sample and negative sample farther away from each other, suggesting that this feedback type is more reliable in depicting user preferences. We will discuss more with experimental results in Subsection \ref{subsec:hyper}.

\section{Experiments}
\label{sec:experiments}
\subsection{Datasets}

The real-world datasets with multiple user feedback types used for experiments are Tmall\footnote{https://www.tmall.com}, Xing\footnote{http://www.recsyschallenge.com/2017/}, and Dianping\footnote{http://www.dianping.com/} corresponding to e-commerce, job, and local services recommendation scenarios, respectively. Interactions including \textit{bookmark}, \textit{reply}, and \textit{response} from a recruiter in Xing are regarded equally as positive feedback type to reduce sparsity. For Dianping, ratings on aspects of \textit{overall}, \textit{taste}, \textit{environment}, and \textit{service} are converted into implicit form. The feedback termed as \textit{visited} denotes a user posted a review on an item without rating it. Users with less than 12 and items with less than 16 interactions in Tmall and Dianping are filtered. While for Xing, the threshold is set to 5 for both user and item. Table \ref{tab:data} summarises the statistics of the filtered datasets. 

\begin{table}[h]
	\centering
	\renewcommand\arraystretch{1.1}
	\footnotesize
	\begin{tabular}{|c|c||c|c|c|p{0.85cm}|}
		\hline 
		Dataset&Type&\#User&\#Item&\#Inter&Density\\
		\hline
		\hline  
		\multirow{4}*{Tmall}&Purchase$^*$ & 84k & 36k & 1.1m & 0.04\% \\
		\cline{2-6}
		&Cart & 3.7k & 5.4k & 7.6k & 0.01\% \\
		\cline{2-6}
		&Collect & 53k & 34k & 582k & 0.02\% \\
		\cline{2-6}
		&Click & 84k & 36k & 9.2m & 0.30\% \\
		\hline 
		\multirow{3}*{Xing}&Positive$^*$ & 19k & 9.6k & 201k & 0.11\% \\
		\cline{2-6}
		&Click & 18k & 9.2k & 272k & 0.15\% \\
		\cline{2-6}
		&Hide & 2.0k & 5.2k & 33k & 0.02\% \\
		\hline 
		\multirow{5}*{Dianping}	&Overall$^*$ & 129k & 27k & 2.2m & 0.06\% \\
		\cline{2-6}
		&Taste & 60k & 12k & 214k & 0.01\% \\
		\cline{2-6}
		&Environment & 53k & 14k & 169k & 0.01\% \\
		\cline{2-6}
		&Service & 57k & 15k & 175k & 0.01\% \\
		\cline{2-6}
		&Visited & 129k & 27k & 4.6m & 0.13\% \\
		\hline 
	\end{tabular}
	\captionsetup{margin=0cm}
	\caption{Statistics of the datasets. k indicates thousand; m indicates million; and \textit{Inter} indicates interaction. '*' means primary feedback.}
	\label{tab:data}
\end{table}

\begin{table*}[]
	\centering
	\renewcommand\arraystretch{1.1}
	\footnotesize
	\begin{tabular}{|c|c||c|c||c|c||c|c|}
		\hline 
		\multicolumn{2}{|c||}{Methods}&\multicolumn{2}{c||}{Tmall}&\multicolumn{2}{c||}{Xing}&\multicolumn{2}{c|}{Dianping}\\[0.75 ex]
		\hline 
		Type&Name&HR@10&NDCG@10&HR@10&NDCG@10&HR@10&NDCG@10\\
		\hline 
		\hline 
		\multirow{5}*{Primary}&MF-BPR&0.1404&0.0693&0.6203&0.5653&0.5784&0.3321\\
		\cline{2-8}
		&CML&0.3213&0.1763&0.6611&0.6012&0.5965&0.3556\\
		\cline{2-8}
		&CML(ASP)&0.3472&0.2014&0.8556&0.7407&0.6067&0.3548\\
		\cline{2-8}
		&LRML&0.3248&0.1821&0.5858&0.5283&0.5910&\textbf{0.3635}\\
		\cline{2-8}
		&LRML(ASP)&0.4112&0.2430&0.8532&0.7364&0.5831&0.3585\\
		\hline 
		\hline
		\multirow{6}*{Multiple}&TCF&0.3065&0.1776&0.7017&0.6263&0.5273&0.3211\\
		\cline{2-8}
		&MFPR&0.4534&0.2783&0.7840&0.6946&0.5836&0.3579\\
		\cline{2-8}
		&MR-BPR&0.4416&0.2566&0.8548&0.7425&0.5975&0.3477\\
		\cline{2-8}
		&MC-BPR&0.3457&0.1931&0.7204&0.6405&0.5348&0.3302\\
		\cline{2-8}
		&VALS&0.3948&0.2486&0.7550&0.6842&0.5233&0.3329\\
		\cline{2-8}
		&MRMN&\textbf{0.5063}&\textbf{0.3042}&\textbf{0.8604}&\textbf{0.7443}&\textbf{0.6132}&0.3614\\
		\hline 
	\end{tabular}
	\captionsetup{margin=0cm}
	\caption{Experimental results on the datasets. The best performance is in boldface. \textsl{ASP} means we treat all types of (positive) feedbacks as primary feedback to dispel doubts on the data size inequality caused by involving more feedback types for a fair comparison.}
	\label{tab:overall_allmodels}
\end{table*}

\begin{figure*}
	\setlength{\abovecaptionskip}{0.1 cm}
	\setlength{\belowcaptionskip}{0 pt}
	\begin{subfigure}[b]{.247\linewidth}
		\includegraphics[width=\linewidth]{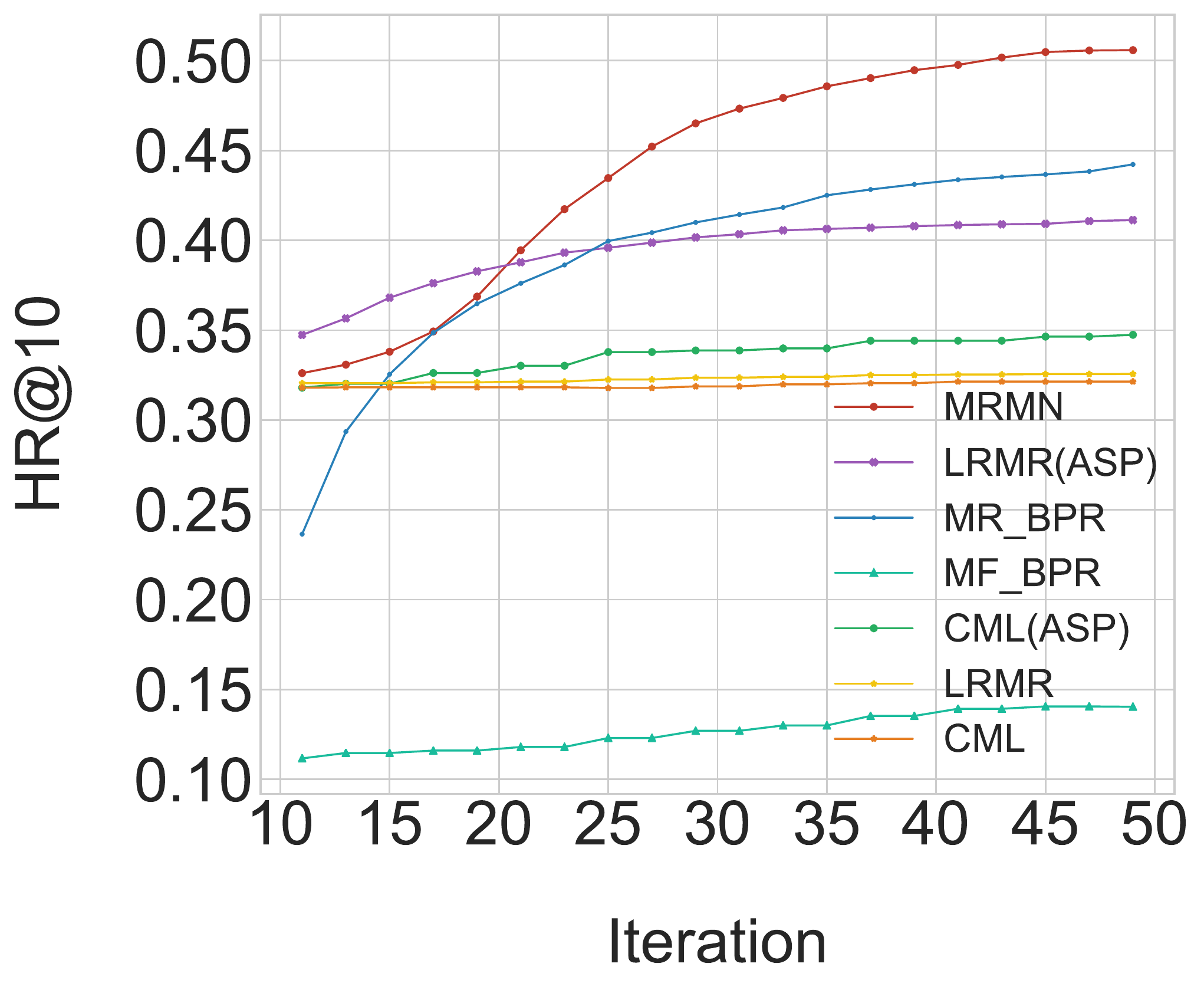}
		\caption{Tmall--HR@10}\label{fig:Tmal_HR}
	\end{subfigure}
	\begin{subfigure}[b]{.247\linewidth}
		\includegraphics[width=\linewidth]{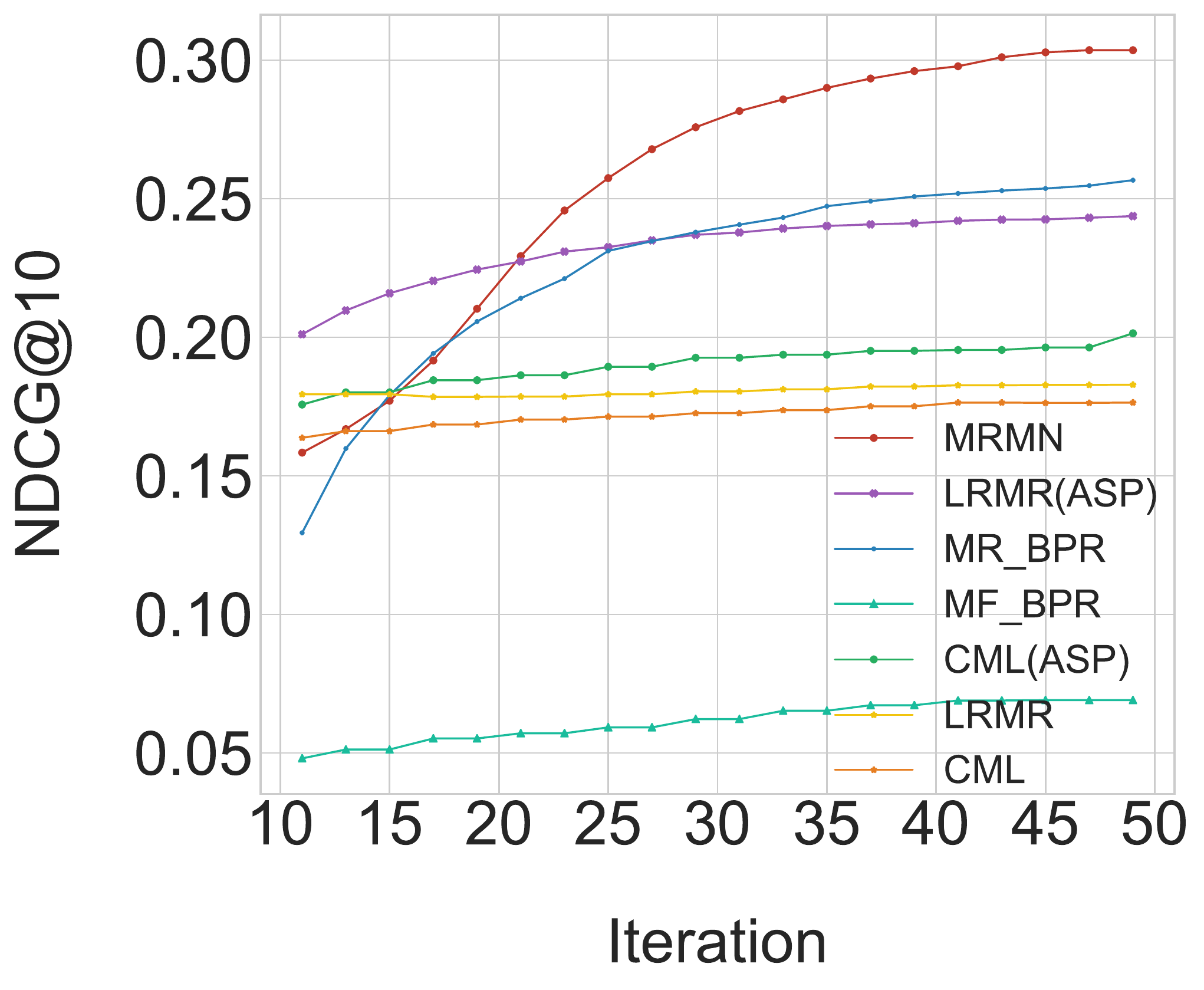}
		\caption{Tmall--NDCG@10}\label{fig:Tmall_NDCG}
	\end{subfigure}
	\begin{subfigure}[b]{.247\linewidth}
		\includegraphics[width=\linewidth]{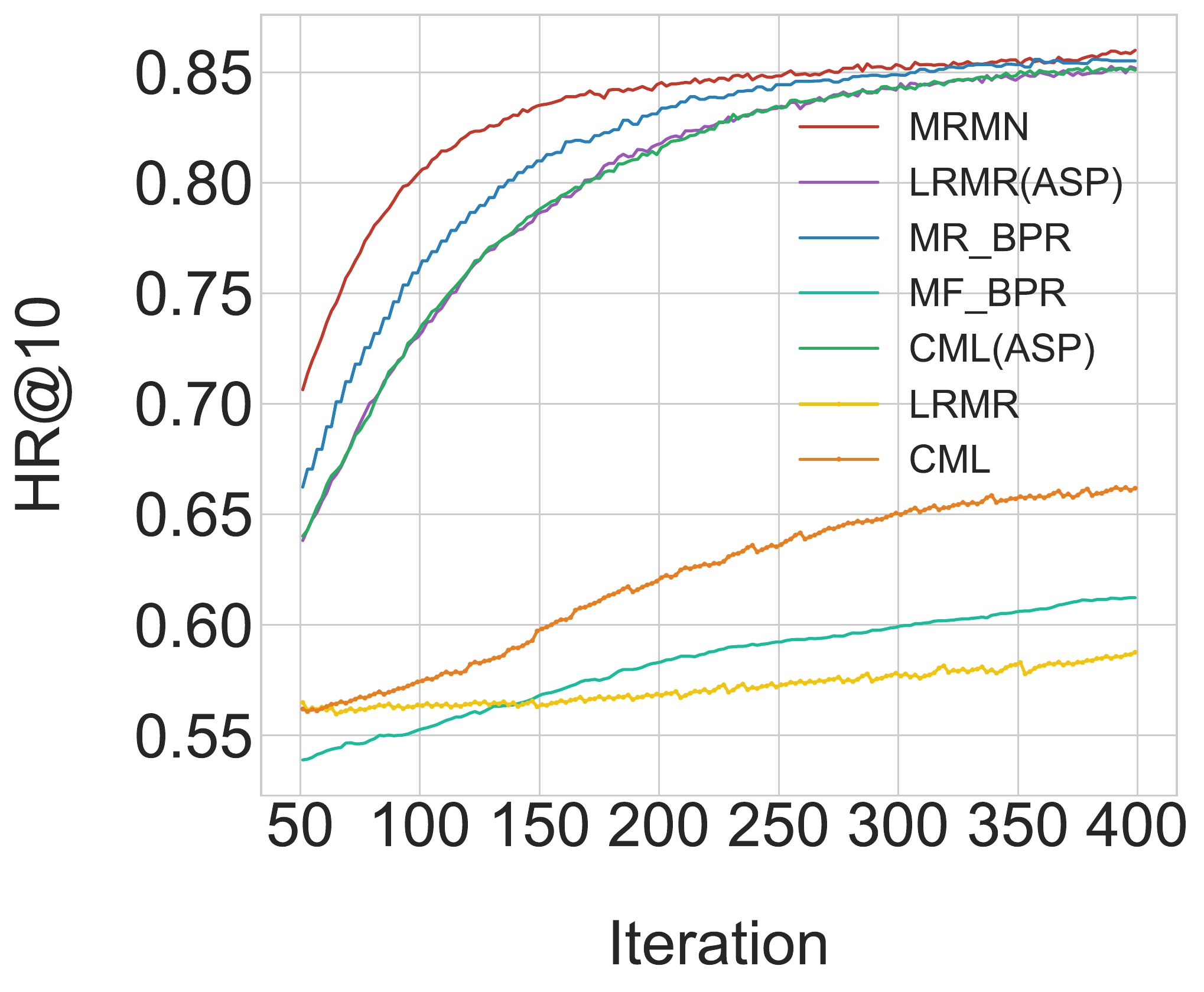}
		\caption{Xing--HR@10}\label{fig:Xing_HR}
	\end{subfigure}
	\begin{subfigure}[b]{.247\linewidth}
		\includegraphics[width=\linewidth]{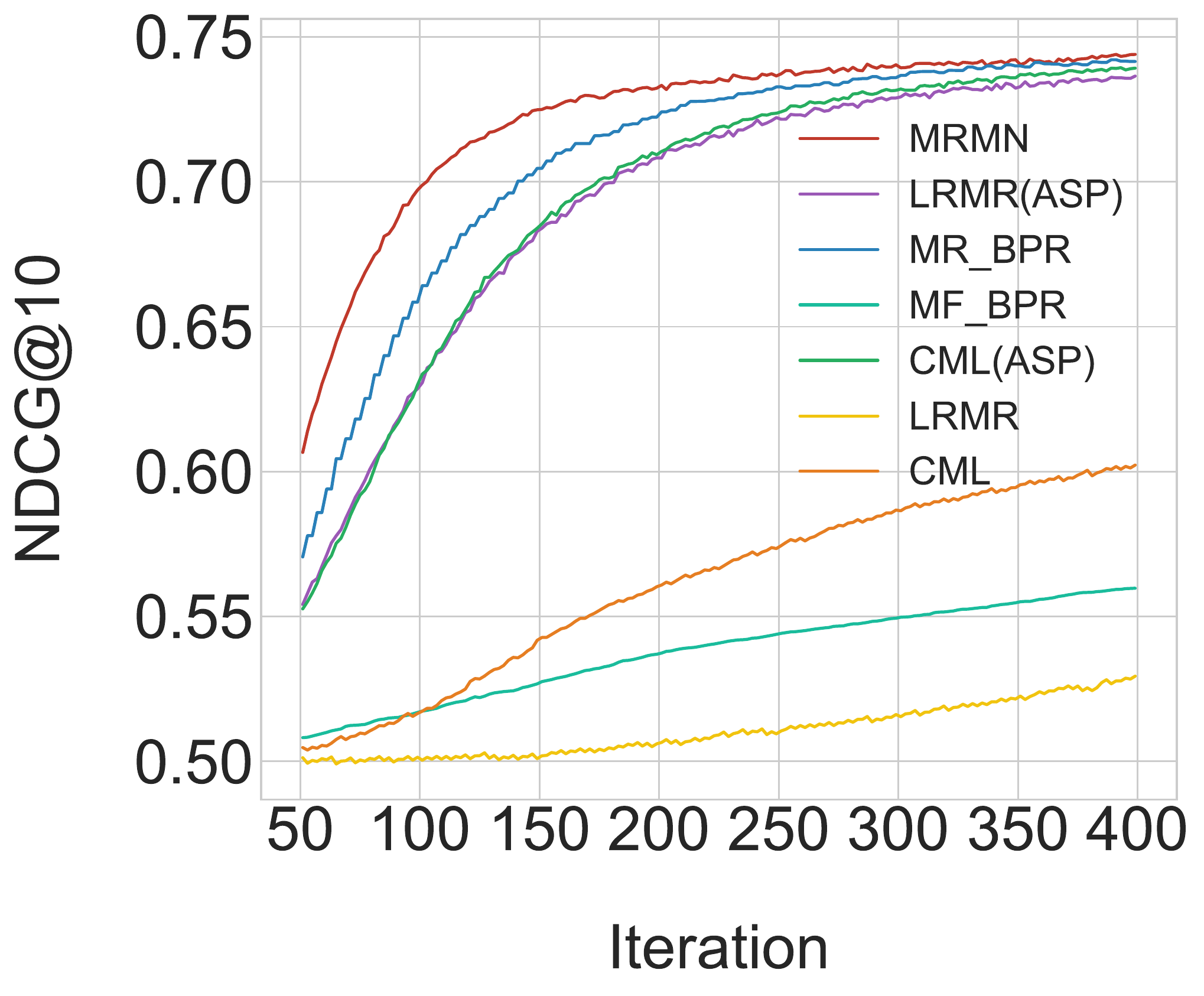}
		\caption{Xing--NDCG@10}\label{fig:Xing_NDCG}
	\end{subfigure}
	\caption{Performance of MRMN and selected baselines w.r.t. the number of iterations on Tmall and Xing.}
	\label{fig:Iteration}
\end{figure*}

\subsection{Evaluation Methodology}
We adopt the \textit{leave-one-out} protocol for model evaluation. Each user's interactions with items are firstly sorted by the timestamps ascendingly. Then the last two records are held out as validation and test set, while the remaining serve as training data. We randomly sample 100 items that the user has not interacted with as negative samples. There is an additional constraint for Dianping that negative samples are selected randomly from places within 3 miles from the positive one. For the two evaluation metrics, \textit{Hit Ratio at K (HR@K)} only considers whether the ground truth is ranked among the top $K$ items; while \textit{Normalised Discounted Cumulative Gain at K (NDCG@K)} is a position-aware ranking metric.

\subsection{Baselines}
According to whether considering additional types of user feedback, baselines are split into two groups. \textit{Primary} group that only employ primary feedback include: 

\begin{itemize}
	\item \textbf{MF-BPR} \cite{rendle2009bpr} It is the MF model using a pairwise ranking loss for optimisation. This algorithm is tailored to implicit feedback recommendations.
	
	\item \textbf{CML} \cite{hsieh2017collaborative} It is a metric learning approach minimising the Euclidean distance between the user vector and item vector.
	
	\item \textbf{LRML} \cite{tay2018latent} This is a metric learning technique employing a memory network to learn the latent relationship between the user and item.
\end{itemize}

\noindent \textit{Multiple} group dealing with multiple feedback types include:
\begin{itemize}
	\item \textbf{TCF} \cite{pan2010transfer} This is a transfer learning approach that is capable of transferring knowledge from auxiliary feedback to the primary feedback. 
	
	\item \textbf{MFPR} \cite{liu2017personalized} It is a method that integrates multiple implicit feedback types in the SVD++ manner. 
	
	\item \textbf{MR-BPR} \cite{ajit2008cmf} BPR framework is applied to collective matrix factorization to make it capable of handling multiple types of implicit data. 
	
	\item \textbf{MC-BPR} \cite{loni2016bayesian} This is a method that samples positive and negative items from multiple relations. 
\end{itemize}

\subsection{Experimental Results}
The experimental results of MRMN along with the baselines for \textit{HR@10} and \textit{NDCG@10} are shown in Table~\ref{tab:overall_allmodels}. As can be observed, MRMN obtains the best performance across all the datasets and metrics in general. In Figure \ref{fig:Iteration}, we also present the recommendation performance of MRMN and six selected baselines of each iteration on \textit{Tmall} and \textit{Xing}. Results on \textit{Dianping} show the same trend and are omitted due to space limitations. In the figure, with the increase of iteration times, the performance of MRMN is gradually improved until convergence is reached. The most effective updates occur in the first 30 iterations and 150 iterations for \textit{Tmall} and \textit{Xing}, respectively. As the architecture of MRMN fuses memory module and multi-relational metric learning, the following will provide a detailed breakdown of our experimental results. 


\subsubsection{Single Feedback against Multiple Feedback Types}

As it shows in Table \ref{tab:overall_allmodels}, multi-feedback methods as a whole provide clearly better performance than single-feedback approaches, especially on datasets of \textit{Tmall} and \textit{Xing}. This is probably due to the fact that \textit{Tmall} and \textit{Xing} consist of binary implicit feedbacks in multiple types, while \textit{Dianping} reflects more about whether a user likes an item in various aspects. Among multi-feedback algorithms, our proposed MRMN performs best that it outperforms the best baseline by about 5\% on \textit{Tmall}, and around 1\% on the other two datasets in terms of the \textit{HR} metric. This phenomenon reveals the effectiveness of integrating multiple types of user feedback in recommender systems.

\subsubsection{Contribution of Relational Metric Learning}

The models applying latent relational learning are MRMN and LRML. Compared with CML, LRML learns a single transition vector for each pair of user and item vectors, while MRMN enables multi-relational learning to identify much more complex interactions existing between them. As we can see from Table~\ref{tab:overall_allmodels} that LRML offers slightly better performance than CML, and MRML further boosts the performance significantly. This observation is particularly evident on \textit{Tmall}, where the improvement brought by multi-relational metric learning is around 10\% on both \textit{HR} and \textit{NDCG}. These findings reveal that the application of multi-relational learning does enhance the recommendation performance since MRML outperforms both of its single-relational counterpart LRML and non-relational learning version CML.  

\subsubsection{Contribution of Memory Mechanism}
The high performances provided by memory augmented networks of MRML and LRML also illustrate the effectiveness of the memory mechanism in recommendation tasks. Moreover, we can detect a hint of memory mechanism's superiority in multi-relational recommendations as the MRMN model performs strikingly better than LRML on almost all the metrics and datasets. 

In a nutshell, the impressive performance of MRMN portrays the successful integration of the memory component and attention mechanism over existing metric learning-based algorithms and multi-feedback models. Additionally, it can be expected to show better performance for a larger dataset with more complex user-item interactions.

\subsection{Hyper-parameter Investigation}
\label{subsec:hyper}
The number of memory slots and key slots $N$ in MRMN is tuned amongst $\{5,10,20,50\}$, and $10$ works best in most cases. The dimension of user and item embeddings $d$ is tuned amongst $\{20, 50, 75, 100\}$ and set to 20 eventually. To explore which combination of $ \lambda ^{\tau }$ on different datasets can provide best performance, and to further investigate the relationships between various types of feedback in a deep sense,  we run experiments with different $ \lambda ^{\tau }$ settings on our datasets and list the results of \textit{Tmall} in Table~\ref{tab:hyperpara} as an example. In the table, the first set of margin values is determined intuitively according to feedback types' importance levels reflected on how likely a user enjoys an item. In the second experiment, the same value is allocated to all the feedback types. In the last row, we reverse the order in the first set, allocating a smaller margin value to a more important feedback type. As the performances of the three settings suggest, assigning higher margin value to more 'important' feedback gives better recommendations on \textit{Tmall}. Similar patterns are also found on \textit{Xing} and \textit{Dianping}. This discussion is not to affirm that setting margin values by experience is always the best. Instead, we show how the MRMN model can be applied to detect the underlying relationships between multiple feedback types.

\begin{table}[h]
\centering
\setlength{\abovecaptionskip}{0.2 cm}
\setlength{\belowcaptionskip}{0cm}
\renewcommand\arraystretch{1.1}
\footnotesize
\begin{tabular}{|c||c|c|}
\hline 
[$ \lambda ^{purchase }$, $ \lambda ^{cart}$, $ \lambda ^{collect}$, $ \lambda ^{click}$]&HR@10&NDCG@10\\
\hline
\hline  
[0.2,0.15,0.1,0.05]&0.5063&0.3012\\
\hline 
[0.1,0.1,0.1,0.1]&0.4648&0.2748\\
\hline 
[0.05,0.1,0.15.0.2]&0.3885&0.2240\\
\hline 
\end{tabular}
\captionsetup{margin=0cm}
\caption{Impacts of margin values on MRMN's performance.}
\label{tab:hyperpara}
\end{table}

\subsection{Relation Visualisation}
\begin{figure}
	\centering
	\setlength{\abovecaptionskip}{0.1 cm}
	\setlength{\belowcaptionskip}{0 pt}
	\begin{subfigure}[b]{.285\linewidth}
		\includegraphics[width=\linewidth]{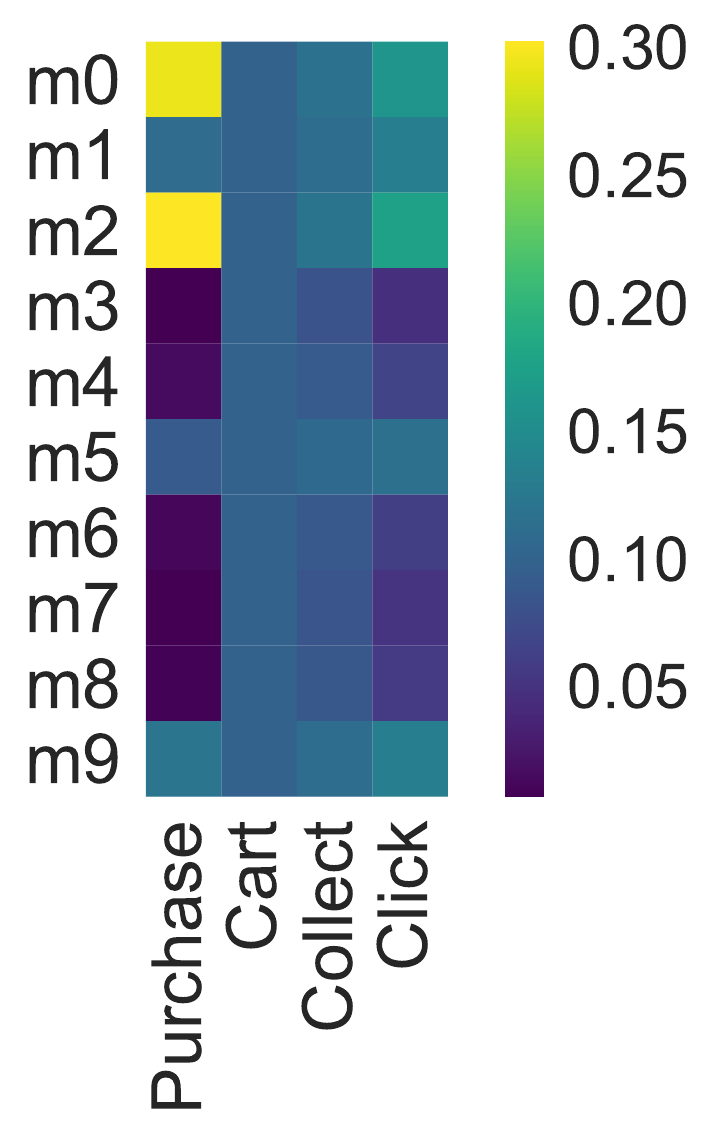}
		\caption{Tmall}\label{fig:Tmall_atten}
	\end{subfigure}
	\begin{subfigure}[b]{.232\linewidth}
		\includegraphics[width=\linewidth]{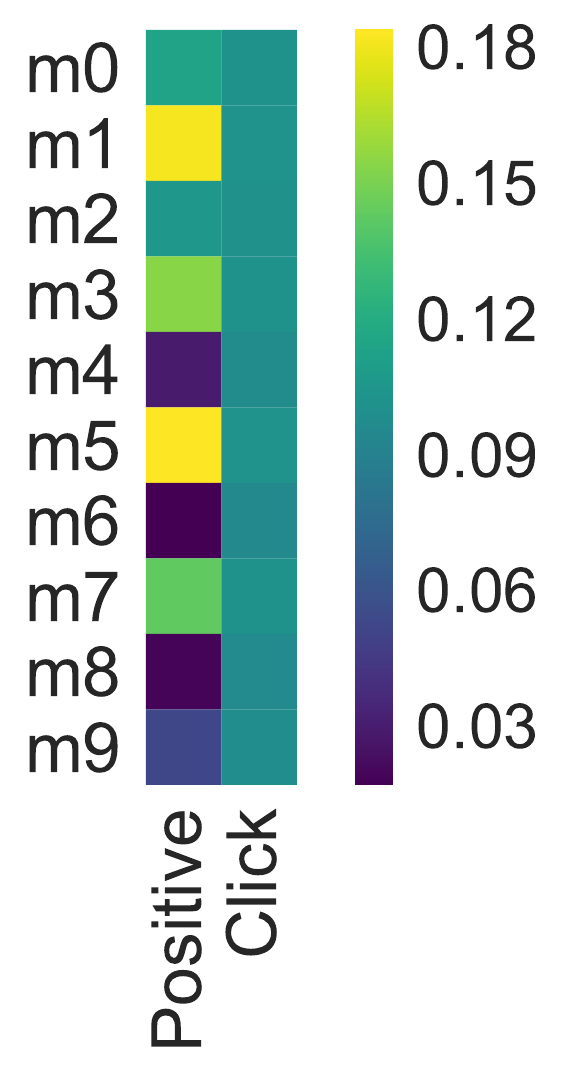}
		\caption{Xing}\label{fig:Xing_atten}
	\end{subfigure}
	\begin{subfigure}[b]{.29\linewidth}
		\includegraphics[width=\linewidth]{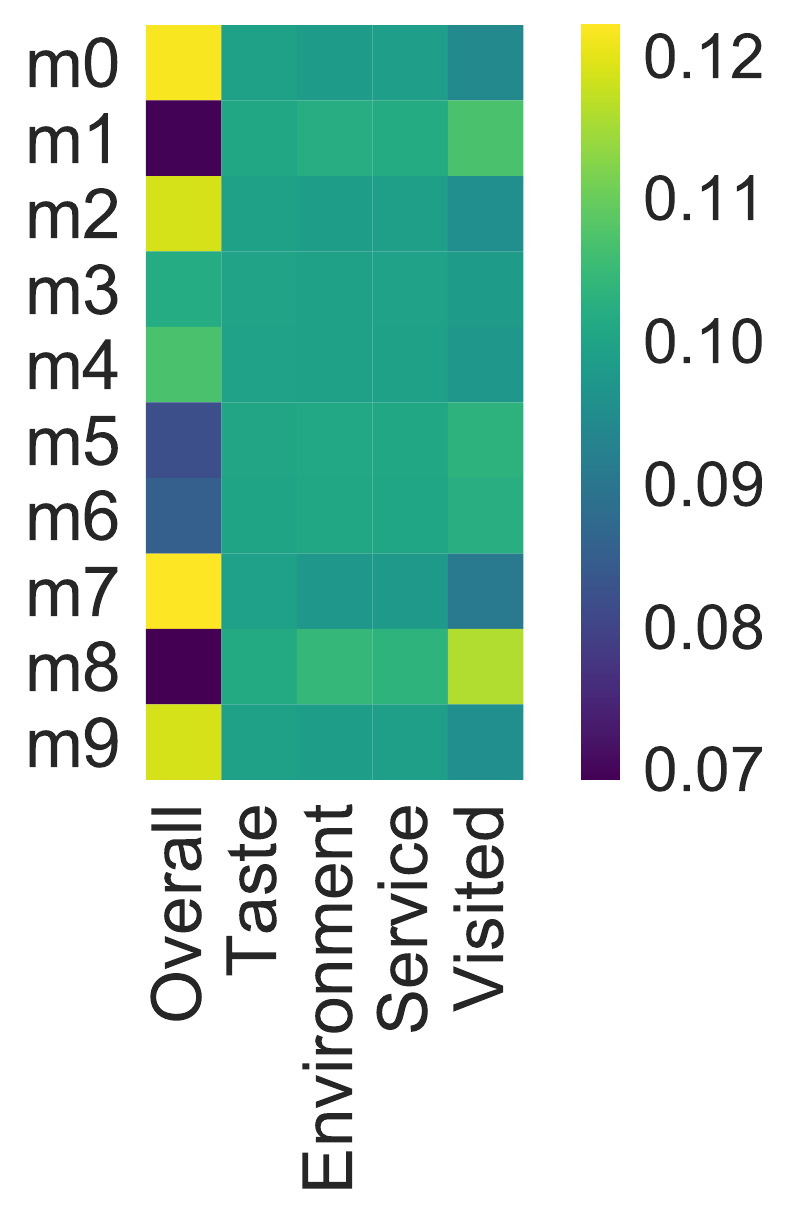}
		\caption{Dianping}\label{fig:DP_atten}
	\end{subfigure}
	\caption{Attention weights over memory slices for feedback types.}
	\label{fig:Atten}
\end{figure}

The attention mechanism in MRMN enables us to visualise the weighted importance of memory slices for multiple feedback types. 
 As can be seen in Figure \ref{fig:Atten}, three datasets present different patterns, indicating that feedback types have different selection rules across memory slices in multi-relational vector learning. Furthermore, the attention weights of popular interaction types are more diverse compared with other minority types. For instance, in Figure \ref*{fig:Tmall_atten}, \textit{Purchase} and \textit{Click} have more diverse attention weights compared with \textit{Cart} and \textit{Collect}. Similar pattern also appears in Figures \ref*{fig:Xing_atten} and \ref{fig:DP_atten}. This finding further proves the necessity of multi-relational detection between each user-item pair that learning an overall relation vector is not fine-grained enough.


\section{Conclusion}
We introduce a novel end-to-end architecture named MRMN, for recommendation with multiple types of user feedback. MRMN is augmented with external memory and neural attention mechanism to capture fine-grained user preference across various interaction space. Comprehensive experiments under multiple configurations demonstrate the proposed architecture's significant improvements over competitive baselines. Qualitative analyses of the attention weights bring insights into the multi-relational learning process and suggest the existence of complex relationships between a pair of user and item, effectively captured by our MRMN model. 

\bibliographystyle{named}
\bibliography{ijcai19}

\end{document}